\documentclass[lettersize,journal]{IEEEtran}
\IEEEoverridecommandlockouts
\usepackage{cite}
\usepackage{array}
\usepackage[caption=false,font=normalsize,labelfont=sf,textfont=sf]{subfig}
\usepackage{stfloats}
\usepackage{url}
\usepackage{verbatim}
\usepackage{amsmath,amssymb,amsfonts}
\usepackage{algorithmic}
\usepackage{graphicx}
\usepackage{textcomp}
\usepackage{xcolor}
\usepackage{tikz}
\usepackage{tikz-3dplot}
\usepackage{tabularx, booktabs}
\newcolumntype{C}{>{\centering\arraybackslash}X}
\usepackage{lipsum} 
\usepackage{caption}

\usepackage{multirow}
\usepackage[flushleft]{threeparttable}

\newcolumntype{b}{X}
\newcolumntype{m}{>{\hsize=.7\hsize}X}
\newcolumntype{s}{>{\hsize=.5\hsize}X}
\newcolumntype{t}{>{\hsize=.3\hsize}X}

\def\BibTeX{{\rm B\kern-.05em{\sc i\kern-.025em b}\kern-.08em
		T\kern-.1667em\lower.7ex\hbox{E}\kern-.125emX}}

\usepackage{listings}
\usepackage{xcolor}

\usepackage{pgfplots}

\colorlet{punct}{red!60!black}
\definecolor{background}{HTML}{EEEEEE}
\definecolor{delim}{RGB}{20,105,176}
\colorlet{numb}{magenta!60!black}

\lstdefinelanguage{json}{
	basicstyle=\footnotesize\ttfamily,
	numbers=left,
	numberstyle=\scriptsize,
	stepnumber=1,
	numbersep=8pt,
	showstringspaces=false,
	breaklines=true,
	tabsize=1,
	frame=lines,
	backgroundcolor=\color{background},
	literate=
	*{0}{{{\color{numb}0}}}{1}
	{1}{{{\color{numb}1}}}{1}
	{2}{{{\color{numb}2}}}{1}
	{3}{{{\color{numb}3}}}{1}
	{4}{{{\color{numb}4}}}{1}
	{5}{{{\color{numb}5}}}{1}
	{6}{{{\color{numb}6}}}{1}
	{7}{{{\color{numb}7}}}{1}
	{8}{{{\color{numb}8}}}{1}
	{9}{{{\color{numb}9}}}{1}
	{:}{{{\color{punct}{:}}}}{1}
	{,}{{{\color{punct}{,}}}}{1}
	{\{}{{{\color{delim}{\{}}}}{1}
	{\}}{{{\color{delim}{\}}}}}{1}
	{[}{{{\color{delim}{[}}}}{1}
	{]}{{{\color{delim}{]}}}}{1},
}

\begin{document}
	
	\title{CSM-H-R: A Context Modeling Framework in Supporting Reasoning Automation for Interoperable Intelligent Systems  and Privacy Protection\\
	}
	
	\author{Songhui Yue, Xiaoyan Hong, and Randy K. Smith}
			
	\maketitle

\begin{abstract}
	The automation of High-Level Context (HLC) reasoning across intelligent systems at scale is imperative because of the unceasing accumulation of contextual data, the trend of the fusion of data from multiple sources (e.g., sensors, intelligent systems), and the intrinsic complexity and dynamism of context-based decision-making processes. To mitigate the challenges posed by these issues, we propose a novel Hierarchical Ontology-State Modeling (HOSM) framework CSM-H-R, which programmatically combines ontologies and states at the modeling phase and runtime phase for attaining the ability to recognize meaningful HLC. It builds on the model of our prior work on the Context State Machine (CSM) engine by incorporating the H (Hierarchy) and R (Relationship and tRansition) dimensions to take care of the dynamic aspects of context. The design of the framework supports the sharing and interoperation of context among intelligent systems and the components for handling CSMs and the management of hierarchy, relationship, and transition. Case studies are developed for IntellElevator and IntellRestaurant, two intelligent applications in a smart campus setting. The prototype implementation of the framework experiments on translating the HLC reasoning into vector and matrix computing and presents the potential of using advanced probabilistic models to reach the next level of automation in integrating intelligent systems; meanwhile, privacy protection support is achieved in the application domain by anonymization through indexing and reducing information correlation. An implementation of the framework is available at https://github.com/songhui01/CSM-H-R.
\end{abstract}


	\section{Introduction}
	\IEEEPARstart{I}ntelligent systems take advantage of context-aware computing in reasoning automation and decision-making in context-rich fields such as smart healthcare \cite{edge-smart-health} \cite{context-aware-healthcare-security}, emergency response \cite{KIANI2023288}, smart cities \cite{9756574},  and smart transportation \cite{9756574}\cite{8910356}. Intelligent systems are thus empowered to adapt to the changing environment, learn from experiences, and often make decisions without explicit human intervention \cite{b1001}. A broad range of technologies involved include rule-based expert systems \cite{bielawski1991intelligent}, probabilistic-model-based \cite{deng2021review}\cite{deepa2021ai}, and modern artificial intelligence (AI) techniques such as deep learning\cite{b1050}, and reinforcement learning \cite{KIANI2023288} \cite{ning2019deep}.
	
	With the advancement of sensor technologies \cite{b1000} \cite{b1000001}, contextual information has evolved from spatial and temporal data to complex facts (the states of various related entities for an application). These facts can usually be calculated from multi-source information (data from sensors deployed for entities) and historical contextual data. The variety and heterogeneity bring significant challenges in the context computing of intelligent systems to fulfill different criteria, such as interoperability, consistency checking, support of sharing and reasoning \cite{b100001} \cite{b100002} \cite{b1001}, privacy protection \cite{context-aware-healthcare-security} \cite{BETTINI2015159}, and requires academia and industry to explore various approaches or a combined approach to handling the context from different domains or applications.
	
\begin{figure}[h]
		\centering
		\includegraphics[width=2.5in]{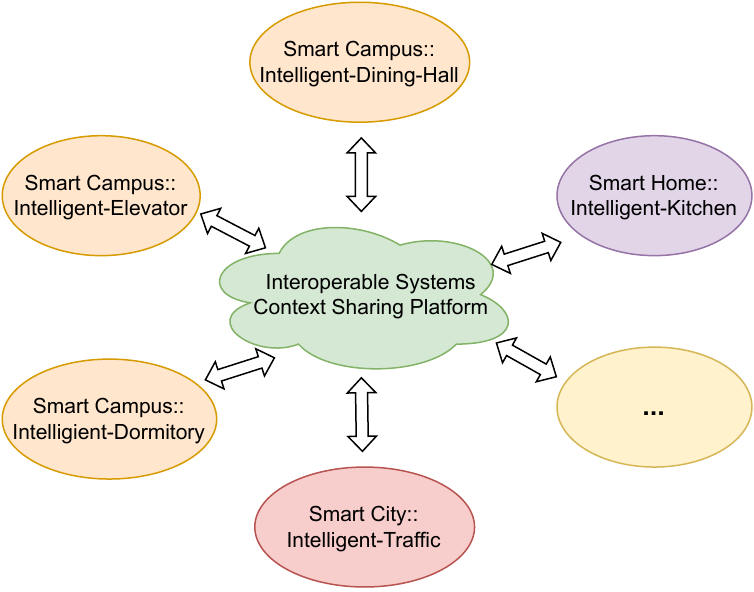}
		\vspace*{3mm}
		\caption{Interoperable Intelligent Systems}
		\label{interoperable_intelligent_systems}
	\end{figure}

	A context modeling framework, CSM-H-R, is proposed in this research to address three challenges. First, as contextual data are multi-sourced, they can be synthesized and then exploited by many intelligent systems simultaneously through proper data sharing and privacy protection mechanisms, with the purpose of reducing the overall data processing effort \cite{DEMATOS2020106988}. This can be achieved particularly with the help of cloud-based architectures and distributed systems \cite{8717579}. Taking a smart campus as an example, applications of Intelligent Elevator \cite{yue2021applying} can be developed. Contextual data collected within a smart campus and those applications can contribute to each other and even to larger areas like a smart city, and vice versa. There could be many intelligent systems co-existing for public services or personal interests within different sizes of areas, and they are interoperable if they work with the same level of contextual information, as illustrated in Figure \ref{interoperable_intelligent_systems}. We refer to the contextual information that can be used directly by those intelligent consumers as High-Level Context (HLC) in this study, in comparison to raw data such as images with RGB pixel matrices or sensor data of accelerometers.
	
	Secondly, context sharing within intelligent systems and semantic interoperability of the shared context \cite{PLIATSIOS2023100754} help reasoning automation in support of decision-making. Especially as data accumulation progresses and the context is prone to changes, when both happen, it is impractical to rely on human experts to identify and embed the explicit and implicit logic within the context in the design of decision-making processes of intelligent systems. Thirdly, privacy protection is one of the key factors in the success of intelligent systems \cite{context-aware-healthcare-security} \cite{BETTINI2015159}\cite{9142779}.
	
	CSM-H-R is based on the ontology model of the static contextual information and Context State Machines (CSM) proposed in our prior research \cite{b1002}, where we apply a state-based approach, particularly in modeling for proactive behaviors and dynamic aspects of context changes \cite{b100201}. Most of the current unimodal context modeling approaches, such as ontology and object-oriented methods, focus on the representation of the current context, like the current values of location (e.g., Home, Office, Gym), timestamp, activity, and the relationship between them \cite{b1000} \cite{b100101}. We proposed that explicitly putting forward states of high-level context can be beneficial in intriguing new angles of context understanding and modeling activities for dynamic state changes and proactive decision-making. Furthermore, the state, as the fundamental element of ontologies’ attributes, also serves as the fundamental element of objects in object-oriented design. Based on these reflections, CSMs were devised to simulate state changes in context attributes and situations.
	
	In summary, CSM-H-R is designed to facilitate reasoning automation through context sharing and interoperating among intelligent systems. The key aspects of our research contribution encompass the following:
	\begin{enumerate}
		\item design of a HOSM framework CSM-H-R, which supports the sharing and interoperation of context among intelligent systems and the components for handling CSMs and the management of hierarchy, relationship, and transition. 
		\item the framework builds on the model of our prior work on the Context State Machine (CSM) engine by incorporating the H (Hierarchy) and R (Relationship and tRansition) dimensions to take care of context dyanmism. 
		\item an implementation of CSM-H-R, providing two designs of input data formats to build core models for different smart campus intelligent systems, demonstrating the sharing of the context within intelligent systems and semantic interoperability of the shared context.
		\item Indexing is applied to reduce the information correlation. Anonymization in transmission and reasoning is discussed in response to privacy concerns.
	\end{enumerate}
	
	The subsequent sections are structured as follows: Section 2 presents the related research. Section 3 outlines the foundational concepts and hypotheses. Section 4 elucidates the methodology of the framework. Section 5 presents the components of CSM-H-R and their interrelationships. Section 6 describes the evaluation of applying the framework. Section 7 is future work and discussion. Lastly, section 8 provides the concluding remarks.
	
	\section{Related Work}
	Hossein et al. \cite{inproceedingsHossein2019} introduce a design of a microservice software framework for implementing automation in the IoT-Fog-Cloud ecosystem for functions from data filtering and transferring to task workflow and scheduling. Elizabeth et al. \cite{contextion2014} present Contextion, a programming framework for creating context-aware mobile applications. The framework is particularly designed for the rapid addition of new sensor technologies on a mobile device. Ralph et al. \cite{bergmann2019procake} presents ProCAKE, a generic framework for building structural and process-oriented case-based reasoning applications. It implements many syntactic and semantic similarity measures for various data types. The framework in this study promotes automation of reasoning, first emphasizing the data collection and transferring data into a format that makes the data more interoperable based on combinational ontology-state modeling so that they can work together to make predictions using various reasoning methods. The implementation of the framework experiments with basic probabilistic/evidence reasoning.
	
	State machine, which is a lightweight, human-readable, and easy-to-parse approach, has been used for activity recognition \cite{b1012}  \cite{b1014}. Teixeira et al. \cite{b1014} present an activity-recognition system for assisted living applications and smart homes, using a lightweight hierarchy of finite state machines (FSMs) to detect actions and activities (sequences of actions). Benitez et al.  \cite{b1012} use a mealy machine to realize action recognition in video sequences. The State Transition Data Fusion Model is proposed by Lambert \cite{b1016}, and it is “a unification of Sensor and Higher-Level Fusion.” The model uses states and transitions to represent real-world status.
	
	Ontology-based context reasoning approaches typically use Resource Description Framework (RDF) and Web Ontology Language (OWL) to build ontology models and then a set of reasoning rules defined to reason on them \cite{meyer2003survey}. Nguyen et al. \cite{van2008context} use a graph-based approach to conduct context reasoning. Machine learning techniques are frequently used for intelligent system-level context reasoning \cite{PORTUGAL2018205} \cite{10054510}. Bousdekis et al. \cite{b1017} utilize the Bayesian Network (BN) for knowledge representation and reasoning under conditions of uncertainty. Mourchid et al. \cite{b1018} present a recommender system for Places Of Interest (POI) using Markov Chain State Models (MCSM). The system leverages contextual information to provide more relevant POI recommendations. Our approach generalizes the usage of states, especially for modeling context, by introducing a state-based conceptual model to systematically model context and promote the computing of context using probabilistic ways such as BN, MCSM, neural networks, or basic CSM matrices.
	
	\begin{figure*}[h]
		\centering
		\includegraphics[width=0.95\textwidth]{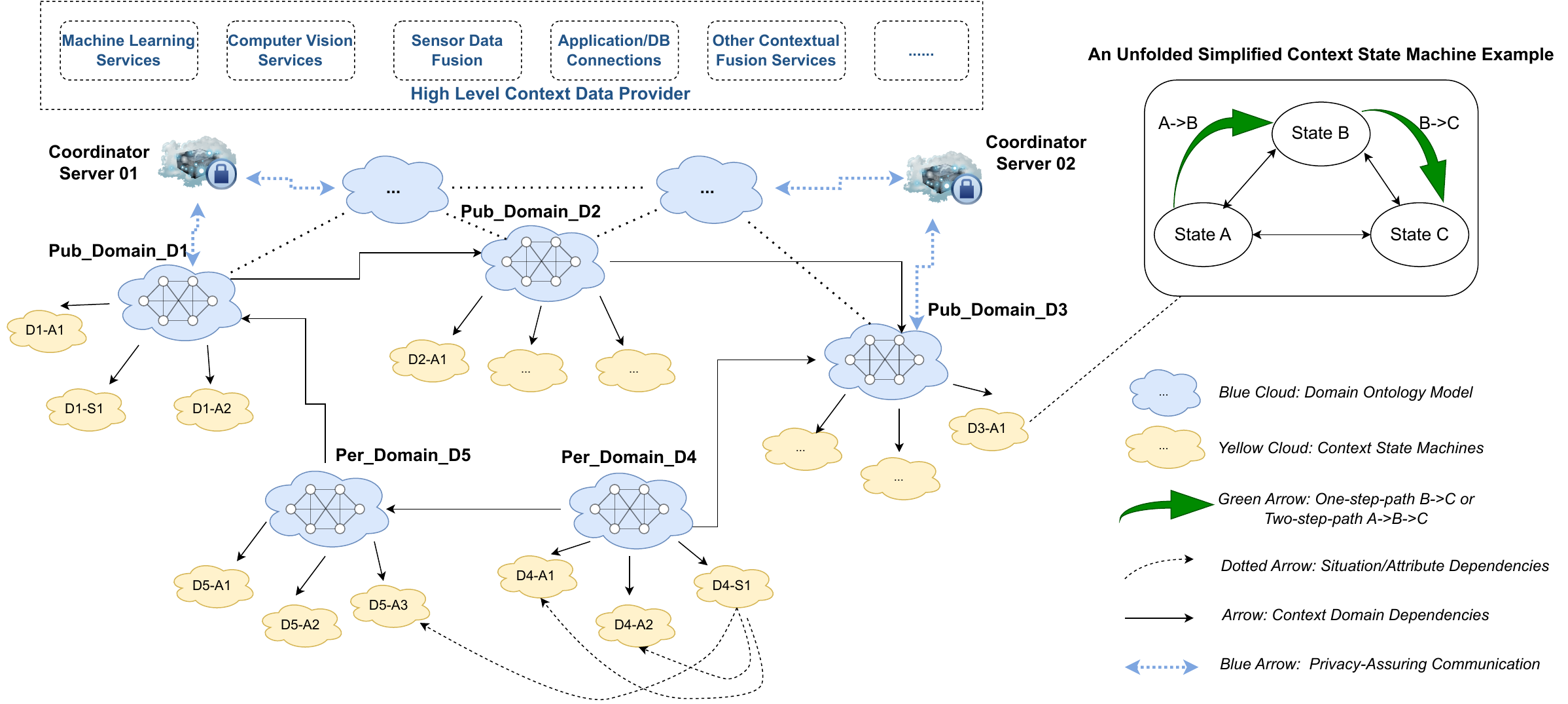}
		\vspace*{3mm}
		\caption{Aerial View of Context Sharing in Conjunction with Modeling Methods}
		\label{aerial}
	\end{figure*}
	
	Ramakrishnan et al. \cite{RAMAKRISHNAN2014207} propose correlation mining algorithms based on Kullback-Leibler (KL) divergence and frequent set mining that exploits correlated contexts to enable unsupervised self-learning. These algorithms help to identify alternate sources for a context and semantically describe the previously unseen contexts in terms of already known contexts. Lee et al. \cite{similarIdea001} propose a modeling method for sensor data in smart space to instantiate and generate a context model and predict the context changes. Compared with Lee's work, our study targets a higher level of contextual data after synthesizing contextual-related information from various sources.
	
	Hoang et al. \cite{9142779} propose a privacy-preserving blockchain-based data sharing platform for the InterPlanetary File System (IPFS), a content-addressable peer-to-peer storage system, to alleviate the concerns also lie in denying of service of centralized systems and the traceability of systems through traditional access control. Mehdi et al. \cite{articleOntologyFrameworkPrivacy2021} propose a three-module framework named “Ontology-Based Privacy-Preserving,” with three modules to store the privacy information using ontologies, find abnormal patterns, and provide a privacy rule manager. Arachchige et al. \cite{arachchige2020trustworthy} introduce a framework named PriModChain that enforces privacy and trustworthiness on IIoT data by amalgamating differential privacy, federated machine learning, Ethereum blockchain, and smart contracts.
	
	\section{Preliminaries}
	This section lays the groundwork for CSM-H-R by providing our observation, research hypothesis, and necessary preliminaries.
	
	\subsection{Observations and Hypotheses}
	Our motivation for this study is based on the following observations and hypotheses.
	
	1) We observed the limitations in the possible numbers of the attributes' states that are used with intelligent system reasoning. Even though for some entities' attributes, the data can be continuous, like the temperature value, a continuous range can be divided into different segments and make the states to be discrete. Some research about working with continuous context can be found in \cite{ji2013cell} \cite{hidasi2014approximate}.
	
	2) A hypothesis is that the number of meaningful states for some problem domains is even smaller and small enough to be computational. For example, after the discretization of temperature values using the rounding method, we can get integers between -20 F and 120 F, which is a possible range for a habitable environment. Considering a specific area, like San Fransisco, the range can be between 32 F to 100 F. After applying proper scaling, we can even only reserve states like "Cold," "Cool," "Moderate," "Warm," and "Hot."
	
	3) For human-centered intelligent systems within IoT, contextual data can change frequently, and so do the decision processes. Thus, contextual reasoning logic needs to adapt to the changes accordingly.
	
	4) Users and designers of intelligent systems sometimes need to understand the context and reasoning rationale. Modeling the state transitions (possibly with different scaling techniques) can help achieve this requirement.
	
	\subsection{Context Modeling and CSMs}
	Context modeling is the core of context processing in supporting context reasoning, context sharing, and semantic interoperability of heterogeneous systems \cite{padovitz2005approach}. Classical context modeling techniques include key-value, markup, logic-based, ontology-based, object-oriented, and graphical modeling. Among them, ontologies are widely regarded as a highly effective solution for knowledge representation and facilitating information interoperability among applications operating in dynamic and diverse environments \cite{PLIATSIOS2023100754}.
	
	Context State Machines (CSMs) are based on the concept of ontologies and promote the states and state transitions. We applied a finite state machine, more specifically, a mealy type state machine, to represent the states' sequences and transitions. The two types of CSMs proposed in one of our prior work are Context Attribute State Machine (CASM) and Context Situation State Machine (CSSM). Figure \ref{aerial} presents an aerial view of context distribution, communication, and modeling for intelligent systems of public domains (with public services related APPs) and personal domains (for applications that are customized for different personal needs).
	
	As shown in Figure \ref{aerial}, each domain can administer contextual information of the entities that belong to the domain, which implicitly contains an ontology model of those entities, and each entity can have different attributes \cite{yue2021applying}. The combinations of the states of some correlated attributes and entities form various situations \cite{b1000}. Each domain can have an attribute like D5-A1, which means the domain D5 has an attribute A1, and a situation like D4-S1, which means domain D4 has a situation space S1. For simplicity, we did not list the entities in between the domains and attributes and the situation spaces in the notations. Additionally, a simplified example of CSM can be found for D3-A1. More examples and details about building them can be found in \cite{b1002}.
	
	The steps representation and coordinator servers in Figure \ref{aerial} will be referred to in the following methodology sections and discussion for the topics of matrix representation and a privacy protection mechanism.
	
	\subsection{HCL Data Examples}
	HCL data can be used as input for intelligent applications or context reasoning middleware. Vaizman et al. \cite{vaizman2017recognizing}  in their research use around ten sensors to collect behavior information such as "Stairs - going up," "Stairs -going down," "Elevator," "Cleaning," and "Singing." Hande et al. \cite{6563930} uses 20 binary sensors to generate 27 different activities. We use the same level of contextual information in this study, and some examples are given in Table \ref{hcl-data-examples}.
	
	\begin{table*}[htbp]
		\begin{threeparttable}
			\caption{HLC Data Examples}
			\label{hcl-data-examples}
			\begin{tabularx}{\textwidth}{sttmmb}
				\toprule
				Object Category  
				& Object & Attribute & State & Conditions/Complement & Sources\\
				\midrule
				Person::Default   & Adam      & BloodSugar          & Low  & Location:::Timestamp & Body Sensors  \\
				Person::Default & Adam        & Action        & Open-CarDoor001    & Location:::Timestamp   & Body Sensors, Car Sensors, GPSs \\
				Person::Default & Adam        & Action        & Driving    & Location:::Timestamp   & Body Sensors, Car Sensors, GPSs \\
				Vehicle::Bus       & Bus\_01       & Action          & Running  & Location:::Timestamp & Accelerometer  \\
				Vehicle::Bus       & Bus\_01       & Speed          & 45mph  & Location:::Timestamp & Accelerometer  \\
				Person::Professor        & Bob       & Action         & Walking   & Location:::Timestamp & Sensor  \\
				Person::Professor        & Bob       & Direction         & South   & Location:::Timestamp & Sensor  \\
				Person::Professor        & Bob       & Speed         & 5mph   & Location:::Timestamp & Sensor  \\
				Person::Student        & Donnie       & Action         & Arrives    & Building003 ::: Timestamp  & GPS, Camera, History Movement Data \\
				Person::Student        & Donnie       & location         & Building003   & Timestamp & GPS, Camera, History Movement Data \\
				Person::Student        & Donnie       & location         & Building003::Corridor-Segment01   & Timestamp & GPS, Camera, History Movement Data \\
				Person::Student        & Donnie       & Action         & TakingElevator  & Complement:::Timestamp & GPS, Camera, History Movement Data \\
				Person::Student        & Donnie       & location         & Building003::classroom012   & Timestamp & GPS, Camera, History Movement Data \\
				Person::Student        & Donnie       & Action         & Leaves Building003   &  Location:::Timestamp & GPS, Camera, History Movement Data  \\
				Person::Student        & Donnie       & Action         & Eating   &  Complement:::Location::: Timestamp & GPS, Camera, History Movement Data  \\
				Person::Student        & Donnie       & Action         & Talking   &  Complement:::Location::: Timestamp & GPS, Camera, History Movement Data  \\
				\bottomrule
			\end{tabularx}
			\begin{tablenotes}
				\item Notes: 1) Double colon used in the Context Category column is to denote the level relations. In the implementation of this paper, the first level is used as the category. The descendants are normally the labels of an object. 2) Triple colon is to denote a parallel relation. 3) A timestamp example: 2023-01-02 15:02:23. 4)  Complement: Take "Eating," for example; its complement can be a food type or a specific food name or object. 5) A complement can refer to another context object in a different domain. For example, the complement of "talking" can refer to a Zoom meeting. Concretely, the meeting's URI can be the unique Zoom ID.
			\end{tablenotes}
		\end{threeparttable}
	\end{table*}
	
	\subsection{Context Dynamism}
	Context dynamism in this study refers to the dynamic aspects of entities, attributes, and states and the corresponding relationship changes within intelligent system domains in the smart city setting. Below are some examples; however, additional instances could exist in real-world scenarios.
	
	1) New Joining of Entity, Attribute, or State, e.g., a new student is joining a smart campus.
	
	2) Relationship Changes, e.g., a student enrolls in a class provided by a professor.
	
	3) Hot-spot Situation Forming, e.g., a student makes a decision only with the accompany of the happening of another fact.
	
	4) Cycles and steps of transitions affecting decision-making: a student will visit a place or make a decision only after a pattern of visiting some other places.
	
	\section{Modeling}
	As a HOSM-based framework, CSM-H-R centers around the core components of concepts in ontology and states. In this section, we present this study's high-level design, including the core class diagram of CSM-H-R, encoding and privacy protection, and leveraging context hierarchy, transition, and relations.
	
	\subsection{a Simplified Class Diagram of CSM-H-R}
	According to the advocator of the object-oriented (OO) approach, the overall context can be viewed as an object \cite{b1040}. In the state-based approach, the universal conceptual atoms include object, attribute, and state, especially attributes are taken out of objects, and states of attributes are taken out of attributes. In this way, it preserves the best extensibility for modeling at both the concept level and programming level.
	
\begin{figure}[htbp]
		\centering
		\includegraphics[width=3.3in]{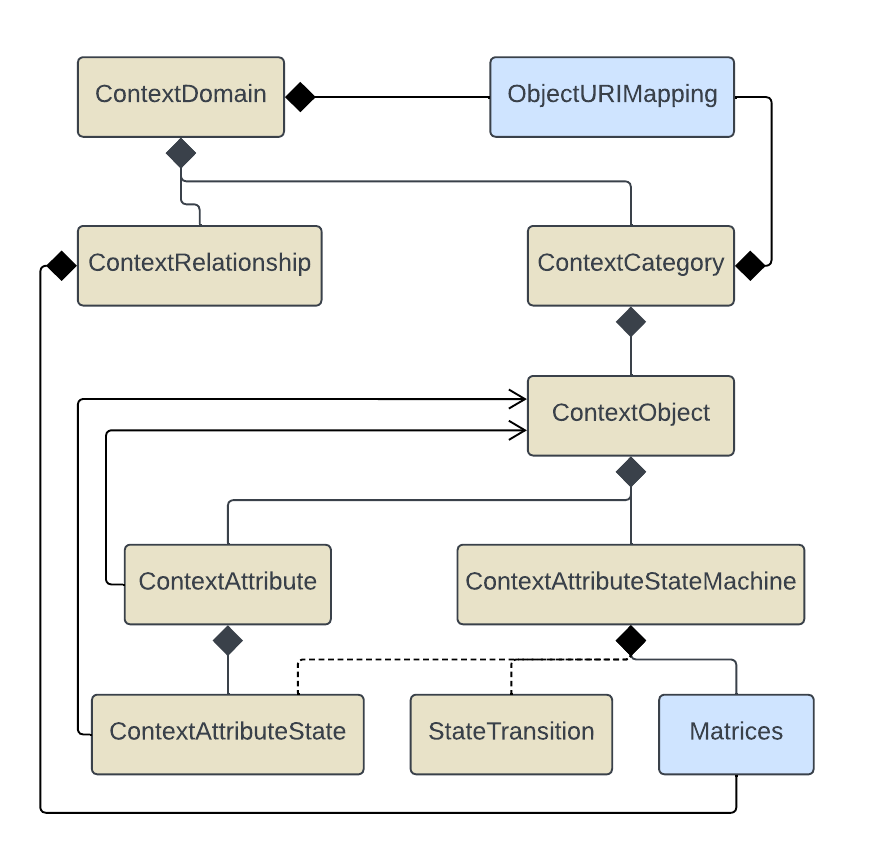}
		\caption{CSM-H-R Core Model (an extension of the CSM core model \cite{yue2021applying})}
		\label{core}
	\end{figure}

	In Figure \ref{core}, the class diagram of the CSM-H-R core model is demonstrated. It extends the CSM core model in our prior work  in three aspects:
	
	1) adding references from "ContextAttribute" and "ContextAttributeState" to "ContextObject" so that it supports the building of hierarchies for context objects (refer to subsection \ref{rh}). 
	
	2) using "Matrices" to represent both the context relationships and state transitions.
	
	3) the utility class of ObjectURIMapping is used to manage the mapping between indexes of objects and the URIs. 
	
	Most of the classes inherit from the class of Ontology by default, and we hide this part for simplicity and readability and only present the composition relations and a reference to represent the hierarchical relation in this class diagram.
	
\begin{figure}[htbp]
		\centering
		\includegraphics[width=3in]{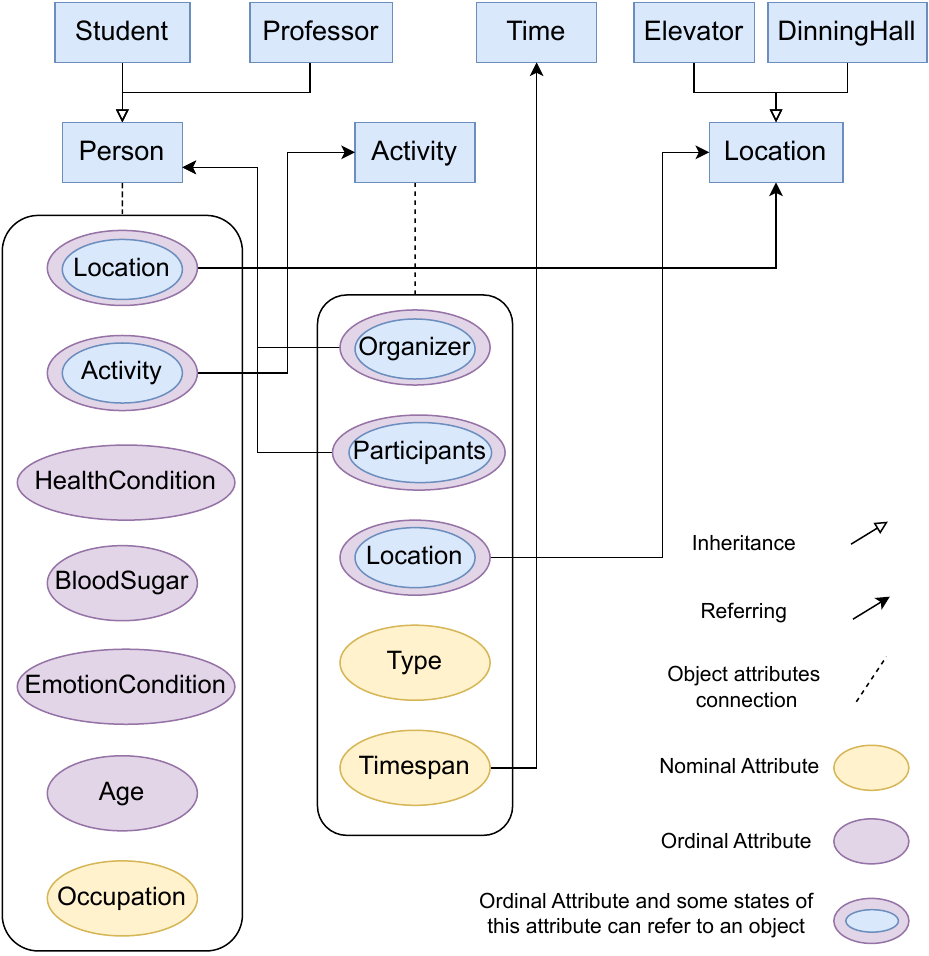}
		\caption{Example of a real domain of objects and attributes}
		\label{exampleOfPersonAttributesAndStates}
	\end{figure}

	There are different context objects in each context category (e.g., a person, a device, or an application). Figure \ref{exampleOfPersonAttributesAndStates} shows a real domain of objects and attributes, where "Person" is a category. "Student" and "Professor" are the occupations mapped to a nominal attribute. Each of the ordinal attributes can be related to a context attribute state machine if necessary.
	
	\subsection{Indexing and Privacy Protection}
	Our current implementation uses indexes to represent the entities in the program and frequency embedding of state transitions in generating multi-dimensional matrices of CSM (refer to the next section of the lower-level design and implementation). To simplify the implementation as proof of the concept, we are using numbers from 0-n as the indexes of the context objects, assuming we have n entities in a context domain. Implementations of other encoding mechanisms can be included in the future.
	
	Privacy protection support is achieved by anonymization through indexing and reducing information correlation. A utility class of "ObjectURIMapping" stores a list of URIs of the entities, and each of them is mapped with its index, which is used in the CSM state transition matrices to represent the corresponding entities. A coordinator server is in charge of storing the URI and the real entity information, as shown in Figure \ref{aerial}. Sensitive data and their owners' information can be shipped to the target computing platforms separately to reduce information correlation. With the help of traditional encryption techniques, the chance of information interception and recovery can be reduced.
	
	\subsection{Context Hierarchy,  Transitions, and Relationships}
	H and R are two special letters of CSM-H-R. The letter H symbolizes the Hierarchies, and R stands for Relationships and tRansitions in CSM multidimensional state transition matrices. They are also treated as two hyperparameters when initializing the domain with the framework: H is how many hierarchies this domain is considering, and R is how many transition steps this domain is considering.  
	
	\subsubsection{Hierarchy}
	Hierarchy has two meanings: one refers to the reference relation between "ContextAttributeState" and "ContextObject" as shown in Figure \ref{core}, and the other refers to different granularities of context attributes in creating CASMs \cite{b1002}. The H in CSM-H-R refers to the first when considered as a hyperparameter.
	
	\subsubsection{Transition}
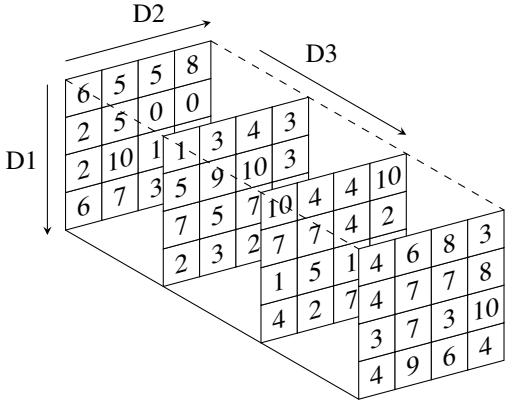
\begin{figure}[htbp]
		\begin{tikzpicture}[x=(15:.5cm), y=(90:.5cm), z=(330:.3cm), >=stealth]
			\draw (0, 0, 0) -- (0, 0, 15) (4, 0, 0) -- (4, 0, 10);
			\foreach \z in {0, 5, 10, 15} \foreach \x in {0,...,3}
			\foreach \y [evaluate={\b=random(0, 10);}] in {0,...,3}
			\filldraw [fill=white] (\x, \y, \z) -- (\x+1, \y, \z) -- (\x+1, \y+1, \z) --
			(\x, \y+1, \z) -- cycle (\x+.5, \y+.5, \z) node [yslant=tan(15)] {\b};
			\draw [dashed] (0, 4, 0) -- (0, 4, 15) (4, 4, 0) -- (4, 4, 15);
			\draw [->] (0, 4.5, 0)  -- (4, 4.5, 0)   node [near end, above left] {D2};
			\draw [->] (-.5, 4, 0)  -- (-.5, 0, 0)   node [midway, left] {D1};
			\draw [<-] (4, 4.5, 10) -- (4, 4.5, 2.5) node [near end, above right] {D3};
		\end{tikzpicture}%
		\caption{A 3-Dimensional CASM Matrix}
		\label{3d-casm-matrix}
	\end{figure}

	A transition refers to a state change, which is also an important aspect of context information \cite{b1002}. Implementing the framework in this study starts with considering 1 transition, and the number of transitions can be 2, 3, or even more, depending on the data analysis and reasoning results evaluation. A two-step transition example A$\rightarrow$B$\rightarrow$C is shown in Figure \ref{aerial}. Figure \ref{3d-casm-matrix} demonstrates the data structure after the embeddings, where the data n in [D1.1, D2.3, D3.4] means the frequency is n to reach a state at index 4 through a state at index 1 and a state at index 3.
	
	CSMs are graph representations of states and transitions, which means there could be circles if the transition number is more than 1. A circle number with its ID can be the next level of transition granularity.
	
	\subsubsection{Relationship}
	A person’s situation could include the friends’, colleagues’, or other related entities’ current states. So, to extract a person’s situation, it is necessary to have a data structure to store the relationship information. In our design, a relationship matrix is used to store this type of information. Firstly, we have an array of persons, and each person has an index that is to be used as the coordinate of the matrix and an array of relationship types. We use a 3-dimensional matrix to store the data. The first and second dimensions are the persons’ indexes, and the third dimension is the relationship types.
	
	For building a person’s CSSMs, the system needs a mechanism to find the current situation and update the situation. The relationship matrix can be used to extract situations and build CSSMs. A situation comprises the person’s state and last state (transition), the person’s related entities’ state, and the last state (transition). In order to obtain the information of the latter, the system will process all triple inputs. For each triple, the system will traverse the relations to see if the triple’s subject has a relationship with the person. If so, the system will add the “subject,” its state, and the latest transition to the situation. For example, we have a person P1 and get the first two features: P1.location.state \& P1.location.transition. We can get the person’s situation using the two features: From the P1.relatinships matrix, we get a list of persons and iterate each of them to get their states and transitions. Then, the combination of their states and transitions composes a situation state.
	
	If there are too many related entities for one entity, to refine the computing process in the future, we could only consider the closest entities’ situations, and the closeness could be defined using the time the entities spend together or other features.
	
	\subsubsection{Relationship and Hierarchy} \label{rh}
	Relationships and hierarchies together can provide richer context than simple text information. Relationship refers to the class relationships or entity relationships in Ontology concepts. It involves the start of a relation (relation registration), updating a relation, and deleting/disabling a relation.
	
\begin{figure}[htbp]
		\centering
		\includegraphics[width=3in]{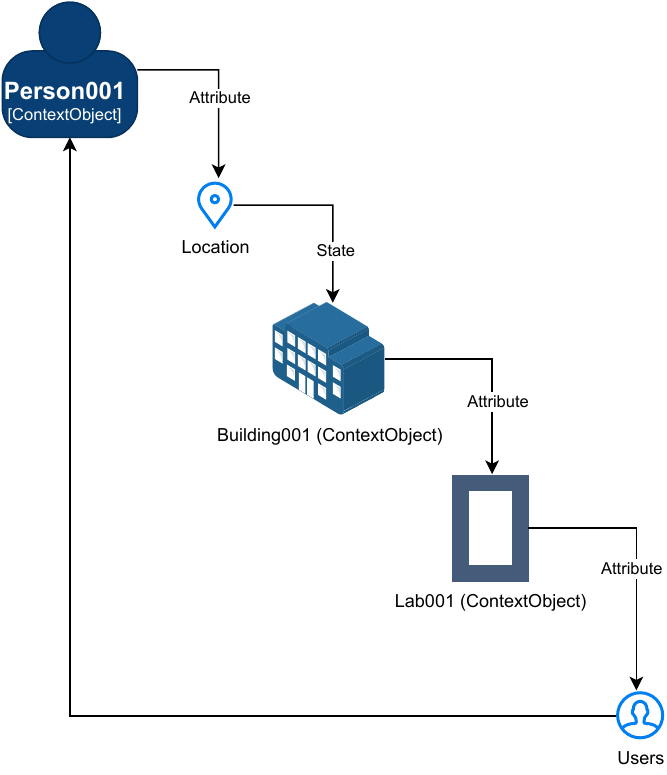}
		\caption{Example of Hierarchy Discovery}
		\label{exampleOfHierarchy}
	\end{figure}

	Relationships can be extracted from relational databases and human inputs or automatically identified through correlations such as co-location and co-timing. Additionally, relationships can also be automatically identified through mining with the hierarchies. An illustration of this process using hierarchical mining is provided below:

	\textit{A person is going to a building; an intelligent system is to reason whether the person will take an elevator. The hierarchy structure is presented in Figure \ref{exampleOfHierarchy}. Suppose the person object is Person001, the attribute is location, the building is Building001 as a state of the location, and Building001 contains a lab called LAB001. According to the CSM-H-R core model, Building001 is an object referred to by the state. LAB001 is an attribute of Building001, and it has a label called owner, or an attribute of a user list, and the label or this list refers to the person Person001. We can see there is a strong bi-directional relationship. So the state of the lab001 will affect the person's decision, then we will give more weights in LAB001 for the person to make related decisions, and further evaluation can adjust those weights to learn a matured model.}
	
	After transferring the contextual data into the objects of an object-oriented programming language, we can use the data stored in the CASMs and CSSMs to formulate more matrices, such as for decisions, to support context reasoning. Contextual data are transferred into numerals in objects and CSMs, so algorithms to reason on the numeral data can be designed based on fuzzy logic, probabilistic logic, or rules.
	
	\section{CSM-H-R: Supporting Context Sharing, Reasoning, and Privacy Protection}
\begin{figure*}[h]
		\centering
		\includegraphics[width=0.95\textwidth]{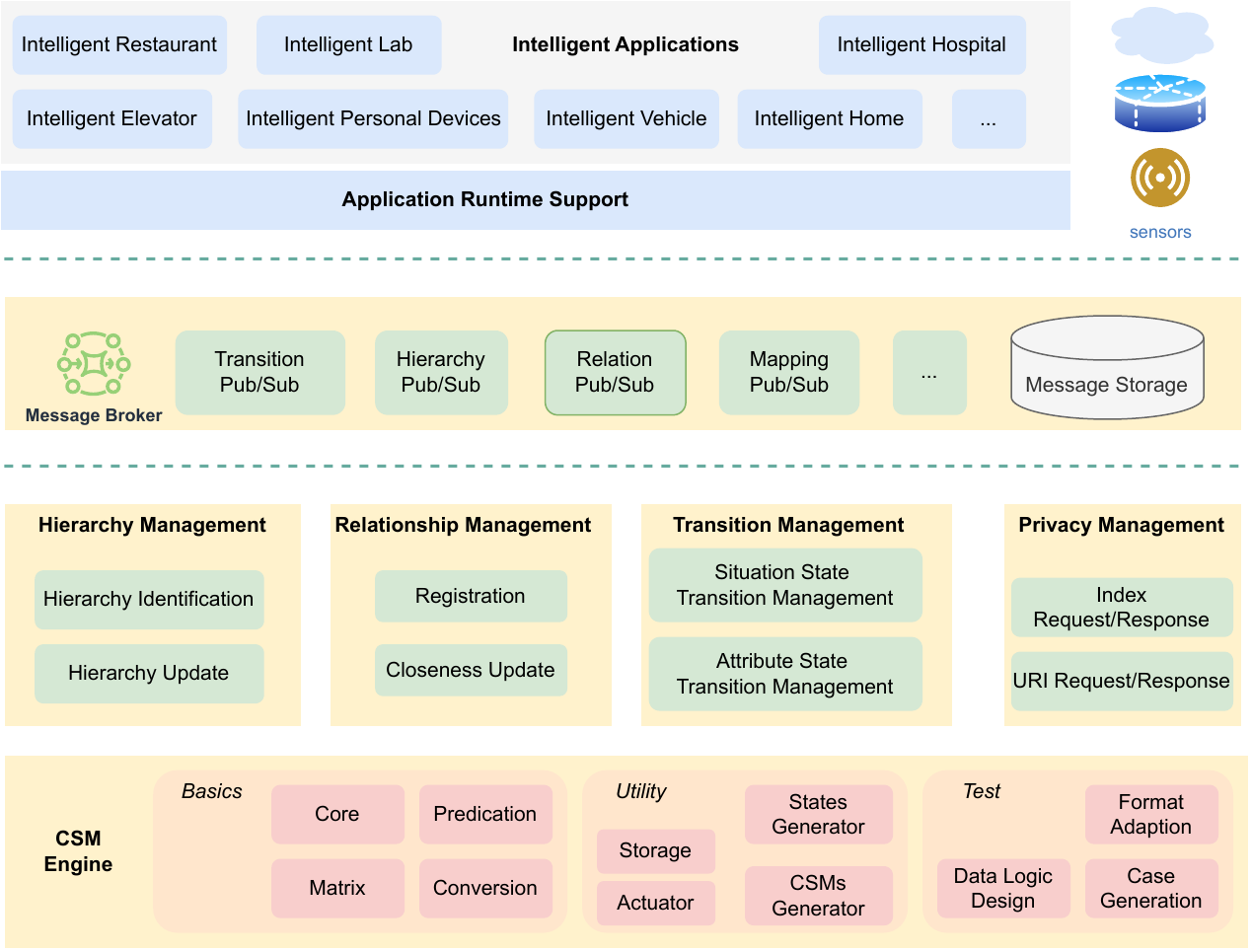}
		\vspace*{3mm}
		\caption{the CSM-H-R Framework for Interoperable Intelligent Systems and Privacy Protection}
		\label{csmhr-framework}
	\end{figure*}

	This section presents the components of CSM-H-R, as shown in Figure \ref{csmhr-framework}. 
	In order to support context sharing and interoperability, the framework is built with a message broker, which consists of channels for context and message tranmission and connecting the intelligent systems and the other context processing components shown in yellow rectangles.
	
	\subsection{CSM Engine}
	The CSM engine was primarily introduced in our prior research \cite{yue2021applying}, and it is designed and developed to let the contextual data live in the states and context state machines. It contains a "core" module, highly related to the state-based modeling of contexts. It defines the basic classes for the concepts in context state machines, including context domain, context category, context object, context attribute, context attribute state, etc. Matrix is the fundamental computing paradigms in this implementation. State transition matrices are used to represent the CASM, CSSM and the relationships. 
	
	To facilitate the data processing of building and configuring CASMs, we implemented two types of triple, as conversion data formats. One is Triple-H-R, a triple of "object," "attribute," and "state," with complements and conditions as shown in Table \ref{hcl-data-examples}. The other is an extension of the Triple concept from the Resource Description Framework (RDF) \cite{yue2021applying}. A Triple in RDF is composed of a subject, a predicate, and an object. Meanwhile, an extended triple Triple-RDF in our framework includes decisions and conditions. The condition can contain several triples to represent the condition details.
	
	The prediction module is designed to contain context reasoning-related classes and attributes such as context reasoning function and threshold. It also includes internal methods to run the CSMs’ context reasoning functions.
	
	\subsection{Utility and Test}
	The utility module contains the code file that can be used to provide utility functions for initialization, generation, storage, and testing. Particularly, the "State Generator" module is designed to load data from various sources. Then, the data can be used for object, attribute, and state extraction. The "CSM Generator" module uses objects, attributes, and states' information to build CSMs for context objects. The "Actuator" module can receive CSMs from or distribute to other servers. It also contains functions for "file" to "machine" (CSM object) and "machine" to "file" transformation. The "file" can be in XML format or from database tables.
	
	One important task in developing the framework prototype is testing it using mock data, which should be generated according to defined cases and data logic. The data may need to be updated to a format the utility functions can process. 
	
	In our implementation of storage functions, we use JASON files to store the CSM data. An object contains all the data for an entity, including the attributes, the states of those attributes, and the state transition information. Due to its lack of a fixed schema, it aligns well with NoSQL databases. 
	
	\subsection{Hierarchy Management}  
	Hierarchy management requires a submodule to identify the hierarchies. It happens when a new entity joins the ontology graph of an intelligent application, an attribute is updated, or a new state of an attribute emerges. 
	
	The Hierarchy Update submodule is responsible for sending the newly identified ones to other modules that utilize the hierarchy relationship, such as the Relationship Management module. It also receives messages through the broker to collect changes in the hierarchies by calling the hierarchy identification submodule.
	
	\subsection{Relationship Management}
	The relationship types and related entities, such as a friend in a system, are from registration and identification processes; the registration refers to the relations extracted from databases or other sources, and identification refers to an automatic method of relation finding through correlations such as co-location or co-timing (refer to subsection \ref{rh}). At the implementation level, if a new person is encountered in the context, the registration process can register the person in the person array. Similarly, if a new relationship type is discovered in the context, the process can register the relation type in the relation type array.
	
	The closeness of the relations can be updated according to the types of relations, the information from hierarchies, and the transitions. In our implementation, we use a number from 0 to 100 to represent the closeness, and when the number is updated to 0, it can mean that the relationship is disabled.
	
	\subsection{Transition Management}  
	Our study categorizes two types of state machines for situation and attribute, and the situation is described by a composition of attributes from one or more objects. The transitions of the states can represent essential changes in the systems. 
	
	State transition identification can include two parts: building a new type of transition and identifying the occurrence of a real transition. New transitions can be added when new states are added to CSMs. When an object's attribute states are changed, the real transition happens.
	
	Other types of transitions involve steps and circles. If a circle emerges, a new type of transition can be created. The meaning of that type of transition may or may not have a special meaning; however, if some special meaning has been identified, then the mapping between the transition type and the meaning will be stored and maintained by this module.
	
	Besides, state transitions are depicted using the matrix representation. When a number in a matrix changes and a threshold of that number has been registered to trigger some action, this module will monitor the messages for that number in the message broker. Thus, one transition may trigger some actions in different intelligent systems.
	
	\subsection{Privacy Protection}
	To protect privacy, domain-specific IDs from indexes enable this framework to hide details of a Person's Identity (PI) when building behavioral patterns. Since all the states are indexed and those indexes are used when making context reasoning, modeled data and sensitive information can be decoupled at the physical level. For example, they can be transmitted through a different HTTP request or other types of requests or even passed through a USB device. Beyond the encryption of transmission data, this separation mechanism adds another level of protection by design if the CSM-H-R is applied.
	
	\subsection{Channels in the Message Broker}  
	We abstract a message broker layer to show the data types of the framework's data exchanges. A message broker may be the best practice, but if the system is lightweight and there is not a demanding requirement for decoupling, scalability, and fault tolerance, an implementation without a broker may also be used. 
	
	Modules in the below components can send messages to each other through the broker to decouple message handling, and they can also receive new context updates from applications, sensors, databases, and other cloud devices.
	
	\subsection{Application Runtime Support}
	This part mainly contains the adaptors. The runtime support can be in the devices that run applications or an independent process of a server serving many applications. If a new transition happens, the applications can take some actions according to the changes in context, possibly by triggering registered call-back functions.
	
	\section{Primary Evaluation of the Framework Implementation}
	We implemented the framework in this study using a general-purpose language. Among the components presented in Figure \ref{csmhr-framework}, we implemented the modeling and some fundamental functionalities, including the CSM Engine, the basics for supporting hierarchy discovery, relationship management, and privacy. We will continue the implementation of other essential functionalities in our future work. The code is maintained at https://github.com/songhui01/CSM-H-R. 
	
	Our prior study showed that the modeling could support context reasoning for system decisions in both basic and complex scenarios \cite{yue2021applying}. The evaluation in this study involves partial core functions supported by the framework. Firstly, CSMs are built for two intelligent systems, namely, IntellElevator (Intelligent Elevator) and IntellRestaurant (Intelligent Restaurant), in a smart campus setting. Secondly, we evaluate the execution time. Thirdly, we compared the compression rate of CSMs with DEFLATE, which is a lossless compression method.
	
	\subsection{Context Modeling for IntellElevator and IntellRestaurant}
	The framework is a context modeling framework. We designed two sets of HLC data for IntellElevator and IntellRestaurant. The two sets use different formats, and the processing logic is designed correspondingly. So currently, our implementation of the framework supports two types of input format (even though the format of input still means that the input should follow a specific protocol). The input of different formats can be transmitted to one unique form in model representation, which can be further used for reasoning as presented in our prior study \cite{yue2021applying}.
	
	\subsubsection{IntellElevator}
	We use a utility script to generate random location changes and decision-making for 50 people in a smart campus setting. Initially, we set a location for each person. An analogy is that 50 people were born within those five buildings, including Home, GYM, DINNINGHALL, COFFEESHOP, and an academic building called the Science and Engineering Complex (SEC).
	
	A number of records can be generated using the format "index, person ID, person name, person type, date, decision, action, action URI, action type, location URI, location name, location type." These records are transformed into tripleRDF records, which will be further processed into CASMs and CSSMs within a context domain object.
	
	\subsubsection{IntellRestaurant}
	When working with a different input format of Triple-H-R, a triple of "object," "attribute," and "state," we use a utility script to generate data in that format. Then, using those generated data records, a context domain object was created along with CASMs, with an emphasis on presenting the enhancement of usability by supporting another format of input. Our future work will connect the models generated by those two formats by carefully designing the correlations among the data.
	
	The data generated for IntellRestaurant follows more complex logic. In one of our test cases, we generated 2000 students and 500 professors as the entities. We use 23 store names and 10 restaurant names as the location states. Besides the location attribute, we also designed six attributes, including health, blood sugar, emotion, age, sex, and occupation. The activity attribute state is generated randomly from five lists of options, e.g., "work activities" contains "Meeting," "Coding," "Planning," "Presenting," and "Researching."
	
	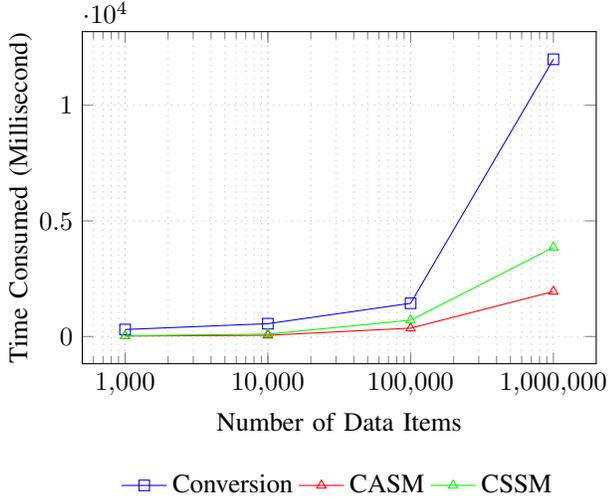
\begin{figure}
		\centering
		\begin{tikzpicture}
			\begin{axis}[
				xlabel={Number of Data Items},
				ylabel={Time Consumed (Millisecond)},
				y label style={at={(axis description cs:0.1,0.5)},anchor=south},
				grid=both,   
				grid style={dotted},  
				xmode=log,
				log ticks with fixed point,
				width=0.95\columnwidth,  
				height=6cm,           
				legend style={at={(0.5,-0.3)},anchor=north, legend columns=3,
					draw=none},  
				]
				
				\addplot[mark=square,blue] coordinates {
					(1000,303)
					(10000,554)
					(100000,1436)
					(1000000,11987)
				};
				
				\addplot[mark=triangle,red] coordinates {
					(1000,20)
					(10000,52)
					(100000,361)
					(1000000,1945)
				};
				
				\addplot[mark=triangle,green] coordinates {
					(1000,29)
					(10000,104)
					(100000,704)
					(1000000,3852)
				};
				\legend{Conversion, CASM, CSSM}
				
			\end{axis}
		\end{tikzpicture}
		\caption{Execution Time for Data Conversion, Building CASM and CSSM for IntellElevator}
		\label{executionTimeForThreePhases}
	\end{figure}

	\begin{figure}
		\centering
		\begin{tikzpicture}
			\begin{axis}[
				xlabel={Number of Data Items},
				ylabel={Time Consumed (Millisecond)},
				y label style={at={(axis description cs:0.1,0.5)},anchor=south},
				grid=both,   
				grid style={dotted},  
				xmode=log,
				log ticks with fixed point,
				width=0.95\columnwidth,  
				height=6cm,           
				legend style={at={(0.5,-0.3)},anchor=north, legend columns=2,
					draw=none},  
				]
				
				\addplot[mark=square,blue] coordinates {
					(10000,52)
					(100000,361)
					(1000000,1945)
				};
				
				\addplot[mark=triangle,red] coordinates {
					(10000,264)
					(100000,1362)
					(1000000,29210)
				};
				
				\legend{IntellElevator, ItellRestaurant}
				
			\end{axis}
		\end{tikzpicture}
		\caption{Comparison of Execution Time for Building CASM}
		\label{ComparisonofTimeForTwoSystemsInBuildingCASM}
	\end{figure}
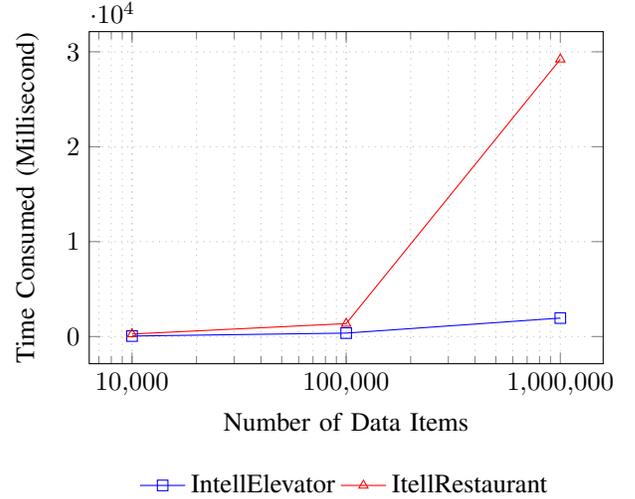

	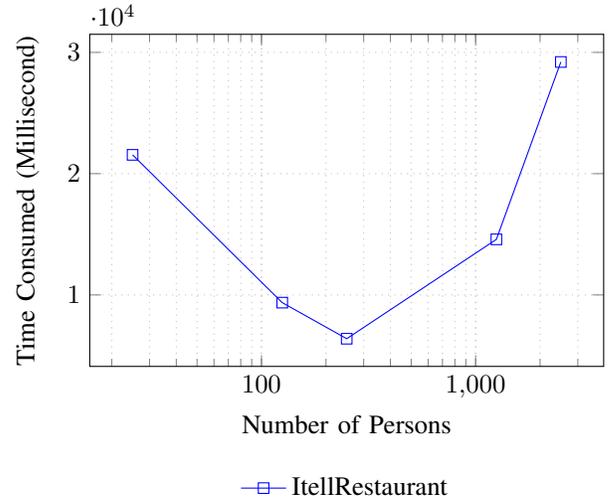
\begin{figure}
		\centering
		\begin{tikzpicture}
			\begin{axis}[
				xlabel={Number of Persons},
				ylabel={Time Consumed (Millisecond)},
				y label style={at={(axis description cs:0.1,0.5)},anchor=south},
				grid=both,   
				grid style={dotted},  
				xmode=log,
				log ticks with fixed point,
				width=0.95\columnwidth,  
				height=6cm,           
				legend style={at={(0.5,-0.3)},anchor=north, legend columns=1,
					draw=none},  
				]
				
				\addplot[mark=square,blue] coordinates {
					(25,21552)
					(125,9365)
					(250,6389)
					(1250,14582)
					(2500,29210)
				};
				
				\legend{ItellRestaurant}
				
			\end{axis}
		\end{tikzpicture}
		\caption{Execution Time for Building CASM for IntellRestaurant w.r.t. Number of Context Objects}
		\label{PersonsNumberAffectExecutionTime}
	\end{figure}
	
	\subsection{Execution time}
	Our prior work proves that the execution time of the modeling and the reasoning process is within a time budget that an intelligent system in literature can tolerate. A Trace-driven simulation was conducted to show that the execution time is short enough for the CPT module to respond to the ICU module of IntellElevator \cite{yue2021applying}. As shown in Figure \ref{executionTimeForThreePhases}, the simulation feeds mocked data into the CSM engine through the Triple-RDF format and uses different amounts of contextual data to generate the time consumption results. 
	
	This study further the examination of the execution by comparing the running time for generating CASMs for IntellElevator and IntellResturant, as shown in Figure \ref{ComparisonofTimeForTwoSystemsInBuildingCASM}. Generating CASMs for IntellRestaurant, when the number of entities changes from 25 to 125, 250, 1250, and 2500, the shape of time used is a convex function, as shown in Figure \ref{PersonsNumberAffectExecutionTime}. The three evaluation figures presented in this subsection apply a logarithmic scale for the x-axis, where it grows slowly at first and faster when the number is larger.
	s
	\begin{table*}[htbp]
		\centering
		\caption{CASM Compression compared with DEFLATE Compression in Processing Data for 2000 Students and 100 professors in the Intelligent Restaurant Application}
		\begin{tabular}{|c|c|c|c|c|}
			\hline
			\multirow{2}{*}{} & \multirow{2}{*}{Input File} & \multicolumn{2}{c|}{Output Files} & \multirow{2}{*}{Ratio (Output Files/Input File)}\\
			\cline{3-4}
			& & CASM Meta File & CASM File &  \\
			\hline
			Uncompressed&55669KB & 5951KB& 1512KB&13.41\%  \\
			\hline
			After Zip (DEFLATE)& 1186KB& 346KB& 239KB& 49.33\% \\
			\hline
			Ratio (ZIP/Uncompressed)& 2.13\%& 5.81\%& 15.81\%&  \\
			\hline
		\end{tabular}
		\label{file-size}
	\end{table*}

	The execution environment was a commercial computer with an “Intel Core i7 CPU @2.60 GHz” using Eclipse with a heap size limit of 8.00 GB. We can gain some valuable insights from execution results:
	\begin{enumerate}
		\item The most relevant finding endorsing the implementation’s fitness for real-world usage is that when the data is within 100k level, the time spent is within 10 seconds in total for the processes of data generation, conversion, and CSMs construction, as shown in Figure \ref{executionTimeForThreePhases}. 
		\item More execution time will be consumed when the entities have more attributes and states to process from comparing the CASMs building for IntellElevator and IntellRestaurant, as shown in \ref{ComparisonofTimeForTwoSystemsInBuildingCASM}. 
		\item When the number of total data items stays the same, the execution time does not necessarily increase linearly when the number of entities increases, as shown in \ref{PersonsNumberAffectExecutionTime}, which suggests the depth of various matrices will affect the efficiency and can be studied for special optimization purposes.
	\end{enumerate}
	
	\subsection{Data Compression}
	Data compression rate can greatly affect the transmission performance in exchanging context information among processes in both local or distributed system settings.
	
	The data compression result compared with the DEFLATE algorithm is shown in Table \ref{file-size}. DEFLATE is a lossless compression method that uses a combination of LZ77 and Huffman coding\cite{venu2022efficient} and is used in formats like ZIP, PNG, and HTTP compression. Our intention is not to directly contrast the compression results with the optimal algorithms in terms of compression ratio. Instead, we aim to provide an overview of how the CSM engine can alleviate the transmission burden by reorganizing the data information. 
	
	CASM meta file and CASM compressed file, as the output of data transformation through CSM Engine, can be used to rebuild the original information. The ratio of raw output-file/input-file is 13.41, which is larger than 2.13, the ratio of ZIP/uncompressed. However, after the compression of the CASM meta file and CASM compressed file using DEFLATE, the total size (346KB + 239KB) is smaller than the size of the file compressed from the input file (1186KB) only using DEFLATE.
	
	\section{Discussion and Future Work}
	Core concepts of the framework and some functionalities are ready to be used as part of the integration and reasoning support for a context-sharing platform or middleware. This paper mainly presents the overall design of the framework based on our proposed CSM-H-R model. We implemented the fundamental modeling and functionalities. For implementation purposes, we plan to enrich the framework by adding more features, such as crafted API and functions supporting hierarchy finding, transition and relationship management, privacy protection, and various encoding methods for fitting machine learning and deep learning algorithms.
	
	\subsection{Strengths and Limitations}
	Besides expressing the essence of context, facilitating context reasoning automation, and supporting privacy protection, our design and implementation of CSM-H-R  in this research demonstrate benefits in four extra specific areas.
	
	Firstly, numerals are used in this research to represent the textual data and object relationships. This way, plain JSON files can be used directly to store contextual data, requiring less storage space than a relational database. This approach can also specifically help machine-to-machine communication \cite{b1050} by minimizing the representation of data: using numerals to represent everything and degrade the data sizes when using different network protocols.
	
	Secondly, the implementation of the CSM engine is designed to be extendable in adding entities and constraints of different system domains for supporting even more complex context-aware scenarios. Taking care of context dynamism is discussed in the next subsection.
	
	Thirdly, data integration from multiple sources can be achieved through first recognizing the entities using ID and URI, and then the data for a specific entiy can be merged or split as required. Related data merging and splitting lgorithms need to be developed in the future.
	
	One disadvantage is due to the vector/matrix representations, and a generated model can lack the semantic details of the data. So, to use a model made of attributes with different granularities, a new data model-building request will need to be initiated. The second disadvantage is the possible states' combinational explosion issue, given the granularity mechanism is not applied, and the meaningful states are not recognized effectively.
	
	\subsection{Taking Care of Context Dynamism}
	The design and implementation of the framework in this study automate the handling of related context dynamism within intelligent systems in the smart city setting. 
	
	1) New Joining of Entity, Attribute, or State: A related object will be created or updated as long as a new one is identified.
	
	2) Relationship Changes. The change of the states will be captured during the processing of the data, and the closeness of the related relationship can be updated.
	
	3) Hot-spot Situation Forming. When a new situation forms, it can be captured in the model for making decisions related to an intelligent system.
	
	4) Cycles and steps of transitions affecting decision-making. Steps of transition are designed as a hyperparameter of the model to reflect the handling of this type of dynamism. Implementation of transition circles is in the future work.
	
	By modeling the dynamic aspects of context and enhancing interoperability, stochastic models and various machine-learning techniques can be adopted. As a result, predictions can be made without designers of intelligent systems to fully master the dynamisms of the entities and events, and the reasoning process only pays attention to the suggestions of the generated mathematical models.
	
	\subsection{CSM-H-R at Scale}
	Although an effective data representation in the IoT area can serve as the backbone of the data for machine learning and data mining techniques usage, one of the concerns of the CSM approach when involving big data is that it can occupy excessive storage and computing resources if there are too many entities and states, or considering extravagant hierarchies and transition steps. However, if the intelligent system is not time-sensitive, the CSMs can be built or updated asynchronously when the server is idler to save computing resources, and the data can be archived after CSMs are built. In addition, the distribution of related tasks to different servers in the cloud can resolve computing bottlenecks and improve responsiveness. The approaches to alleviating the CSM engine server pressure, such as parallel and distributed computing, pends exploration.
	
	\subsection{Explanability}
	We argue that the framework proposed in this study can provide explainability of context reasoning by supporting a gray-box \cite{pintelas2020grey} learning paradigm: on the one hand, through index and frequency embeddings, the reasoning of context has been transformed to numeral computing, and the reasoning engine does not need to know the meaning of the states. Therefore, the usage of numerals helps in the interoperability and sharing of context during context reasoning. On the other hand, the states are mapped to meaningful ontologies, which gives it the potential to reveal the logic of decision-making as one direction of our future work.
	
	\subsection{Generalization}
	Theoretically, CSM-H-R, as a modeling and automatic reasoning approach, can be utilized not only with HLC but also with lower-level knowledge as long as objects, attributes, and states can be recognized. Constraints may vary compared with the HLC-based study in this research and pends exploration.
	
	\section*{Conclusion}
	In order to facilitate the building of interoperable intelligent systems with various context dynamism through context sharing and support privacy protection, a context modeling framework CSM-H-R is proposed in this paper. As an extension of our prior work of applying CSMs to smart elevators, the framework is generalized to support context-rich intelligent systems that require context reasoning and sharing. As presented in this study, the framework's benefits include supporting automatic reasoning, interoperability, context-sharing, privacy protection, and explainability. It takes care of context dynamism and can be applied to context data that is not limited to HLC.
	
	We implemented the framework using a general-purpose programming language. The code is available at https://github.com/songhui01/CSM-H-R. This study presents a primary evaluation of an implementation of CSM-H-R for supporting CSM building using two customized formats (each for one type of intelligent system), execution time, and data compression. The evaluation results of the novel framework demonstrates that CSM-H-R is feasible in modeling context for smart applications, the time efficiency for the data processing, and the facilitation of context sharing through data compression. Additionally, the research discusses the inherent attribute of CSM-H-R related to its support for privacy protection.
	
	\bibliography{mybibtex}
	\bibliographystyle{ieeetr}
	
	\newpage
	
	\begin{IEEEbiography}[{\includegraphics[width=1in,height=1.25in,clip,keepaspectratio]{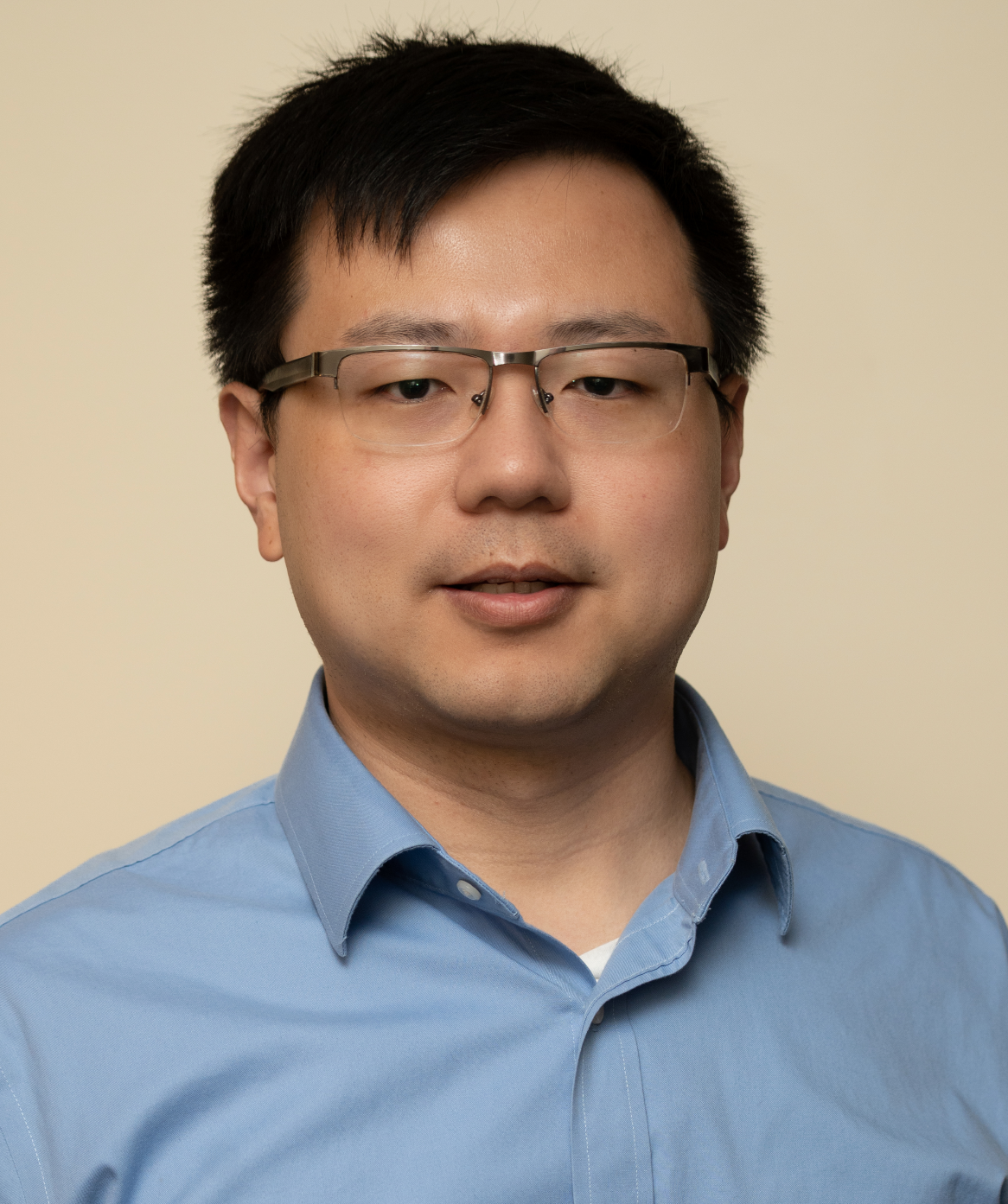}}]{Songhui Yue}
		Dr. Songhui Yue is an assistant professor in the Department of Computer Science at Charleston Southern University, North Charleston, South Carolina. He received a Ph.D. in Computer Science in August 2019 from the Department of Computer Science at the University of Alabama in Tuscaloosa, Alabama. In 2009, Dr. Yue graduated with a B.E. in Software Engineering Institute from East China Normal University in Shanghai, China. Dr. Yue's general research lies in the fields of Smart Software Engineering, Context Reasoning for IoT, Automation of Intelligent Systems, Smart City, Artificial Intelligence, and Machine Learning. Dr. Yue has over five years of industry experience in Web and Mobile-based application development and distributed data processing. 
	\end{IEEEbiography}
	
	\begin{IEEEbiography}[{\includegraphics[width=1in,height=1.25in,clip,keepaspectratio]{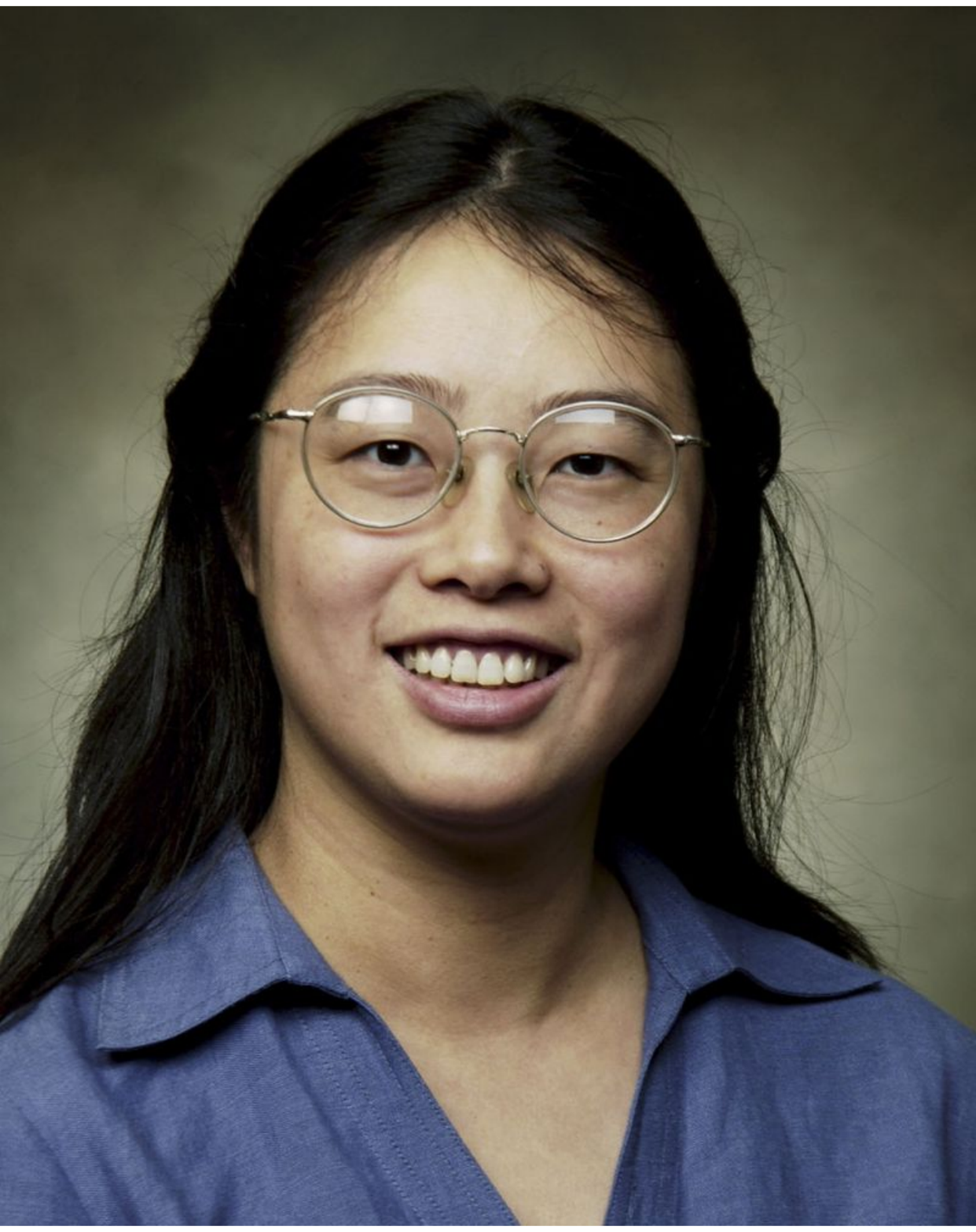}}]{Xiaoyan Hong}
		Dr. Xiaoyan Hong is an Associate Professor in the Department of Computer Science at the University of Alabama. Dr. Hong received a Ph.D. in Computer Science from the University of California at Los Angeles in 2003. Earlier, Dr. Hong obtained a M.S. degree from the Computer Science Department at Zhejiang University, P.R. China. Dr. Hong's research interests include mobile and wireless networks, future wireless Internet, and high-performance networks. Her current research covers mobile ad hoc networks, wireless Mesh networks, vehicular networks, and delay-tolerant networks. Dr. Hong is also involved with a project for the NSF Research Experience for Undergraduates National Academy of Engineering Grand Challenges Program.
	\end{IEEEbiography}

	\begin{IEEEbiography}[{\includegraphics[width=1in,height=1.25in,clip,keepaspectratio]{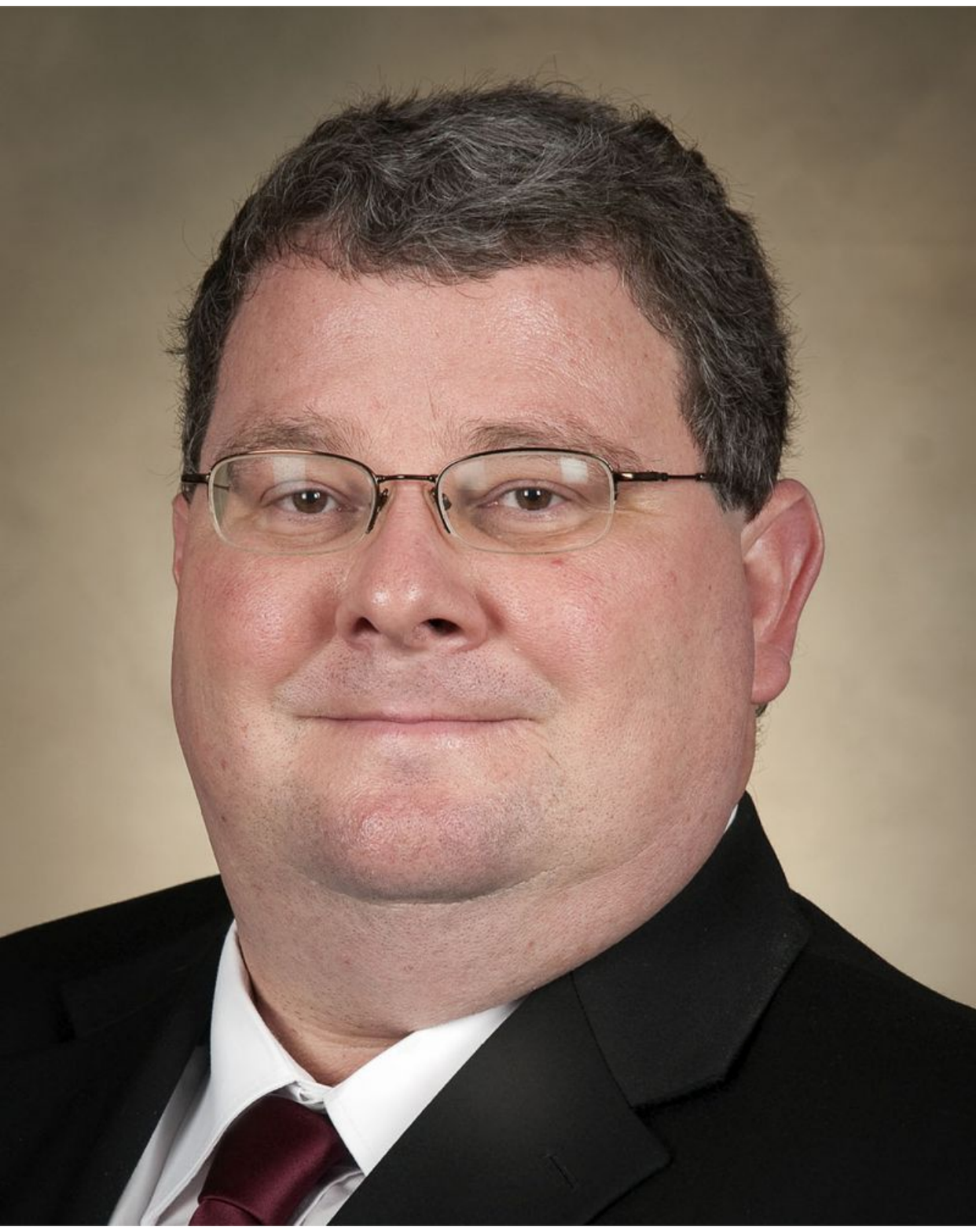}}]{Randy K. Smith}
		Dr. Randy Smith is an Associate Professor in the Department of Computer Science at the University of Alabama. Dr. Smith has over 20 years of experience working at the intersection of Computer Science and Transportation Safety. Dr. Smith has worked with multiple states' Department of Transportation and currently oversees a portfolio of research projects ranging from crash data curation to near real-time digital twins for smart corridors. Dr. Smith is currently serving as the Director of the Center for Transportation Operations, Planning, and Safety at The University of Alabama. He also serves as a principal investigator for the Center for Advanced Public Safety at the university and is a faculty affiliate with the Alabama Transportation Institute.
	\end{IEEEbiography}
	
\end{document}